%% file: main.tex
\documentclass[10pt,twocolumn,letterpaper]{article}

\usepackage[pagenumbers]{cvpr}      
\usepackage{booktabs} 
\usepackage{multirow}
\usepackage{colortbl}
\usepackage[table,xcdraw,dvipsnames]{xcolor}
\usepackage{float}
\usepackage{pifont}
\usepackage{tabularx}
\input{preamble}

\definecolor{cvprblue}{rgb}{0.21,0.49,0.74}
\usepackage[pagebackref,breaklinks,colorlinks,citecolor=cvprblue]{hyperref}

\def\paperID{4134} 
\def\confName{CVPR}
\def\confYear{2024}

\title{Seed Optimization with Frozen Generator for Superior Zero-shot Low-light Enhancement}

\author{Yuxuan Gu\textsuperscript{1},Yi Jin\textsuperscript{1},Ben Wang\textsuperscript{1},Zhixiang Wei\textsuperscript{1},Xiaoxiao Ma\textsuperscript{1},\\Pengyang Ling\textsuperscript{1},Haoxuan Wang\textsuperscript{1},Huaian Chen\textsuperscript{1},and Enhong Chen\textsuperscript{1} \\
\textsuperscript{1}University of Science and Technology of China}
\usepackage{amsmath} 
\DeclareMathOperator*{\minimize}{min} 
\DeclareMathOperator*{\maximize}{max} 
\begin{document}
\maketitle
\input{sec/0_abstract}    
\input{sec/1_intro}

\input{sec/2_related_works}
\input{sec/4_proposed_method}

\input{sec/5_experiments}
\input{sec/6_conclusion}
{
    \small
    \bibliographystyle{ieeenat_fullname}
    \bibliography{main}
}

\end{document}

%% file: preamble.tex
\usepackage[dvipsnames]{xcolor}


%% file: sec/0_abstract.tex
\begin{abstract}
In this work, we observe that the generators, which are pre-trained on massive natural images, inherently hold the promising potential for superior low-light image enhancement against varying scenarios.
Specifically, we embed a pre-trained generator to Retinex model to produce reflectance maps with enhanced detail and vividness, thereby recovering features degraded by low-light conditions.
Taking one step further, we introduce a novel optimization strategy, which backpropagates the gradients to the input seeds rather than the parameters of the low-light enhancement model, thus intactly retaining the generative knowledge learned from natural images and achieving faster convergence speed. 
Benefiting from the pre-trained knowledge and seed-optimization strategy, the low-light enhancement model can significantly regularize the realness and fidelity of the enhanced result, thus rapidly generating high-quality images without training on any low-light dataset.
Extensive experiments on various benchmarks demonstrate the superiority of the proposed method over numerous state-of-the-art methods qualitatively and quantitatively.
\end{abstract}

%% file: sec/1_intro.tex
\section{Introduction}
\label{sec:intro}
$\indent$ Capturing high-quality images under poor lighting conditions is an extremely challenging task since the captured image easily suffers severe visual degradation, such as poor illumination and color distortion. Such degradation significantly affects the performance of the downstream tasks ~~\cite{Degradation_Zheng_2021_ICCV,Degradation_liu2016ssd,Degradation_islam2020semantic}. To address this problem, recent works ~~\cite{Supervised_wu2022uRetinex,Supervised_cai2023Retinexformer,Supervised_xu2022snr,Supervised_jiang2023wavelet} have applied supervised learning to achieve low-light image enhancement (LIE), in which the model is trained with an elaborately collected dataset containing enormous low-light/normal-light image pairs. These methods have achieved remarkable performance, but they have to collect a large number of paired images. Such a data collection operation is tedious and time-consuming. 

To eliminate the requirements of low/normal-light image pairs, many unsupervised learning approaches have been proposed ~~\cite{UnSupervised_jiang2021enlightengan,UnSupervised_guo2020zerodce, UnSupervised_fu2023PairedInstance,UnSupervised_yang2023nerco, UnSupervised_liang2023iterative}. These methods leverage the non-reference low-light images or unpaired low/normal-image images for training, and thus greatly reduce the efforts of data collection. As typical examples, EnlightenGAN ~~\cite{UnSupervised_jiang2021enlightengan} leverages the unpaired images with a generative adversarial network, and Zero-DCE~~\cite{UnSupervised_guo2020zerodce} estimates pixel-wise high-order curves with a set of non-reference losses. These methods can effectively brighten the low-light images. However, the performance of these methods highly depends on the data distribution between the training samples and testing samples. The performance may suffer a dramatic decline when the training and testing samples have a large discrepancy.
\begin{figure}[!t]
    \centering
    \includegraphics[width=\linewidth]{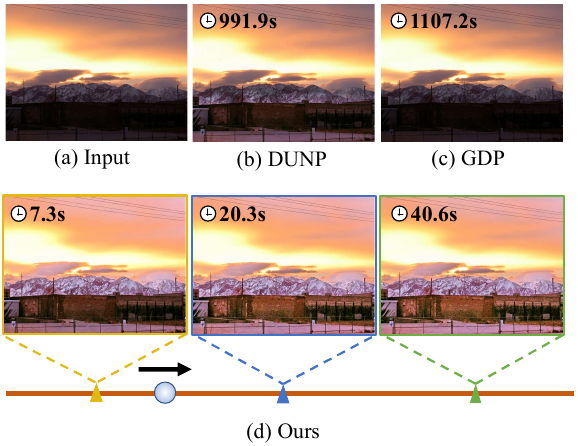}
    \caption{Visual Comparison on a Representative Low-Light Image. All results are obtained under identical computational resources. As iterative optimization proceeds, Our method generates brighter and more visually appealing results in a shorter time frame compared to two other leading zero-shot low-light enhancement approaches. The image outlined in blue represents the output using our recommended iteration number.}
    \label{fig:teaser}
\end{figure}
Subsequently, some methods~\cite{Zeroshot_fei2023gdp, Zeroshot_liang2022dunp} have been devoted to exploring a zero-shot LIE approach that can be trained with a low-light image itself. Representatively, the pioneering work Liang $et\ al.$~\cite{Zeroshot_liang2022dunp}, abbreviated as DUNP, applies the deep image prior (DIP)~\cite{ulyanov2018dip} to LIE tasks, which successfully brightens the low-light images. However, DIP utilizes randomly initialized weights, relying solely on information from a single low-light image. Consequently, it struggles to accurately decompose images, especially in extremely low-light cases. 

In this work, we observe that the generators, which are pre-trained on massive natural images, inherently hold the promising potential for superior low-light image enhancement against varying scenarios. Based on this observation, we propose a novel zero-shot LIE perspective, $i.e.$, strengthening Retinex decomposition with a well-trained generator, to leverage abundant pre-trained knowledge for low-light image enhancement from a single image. Such a perspective takes advantage of high-quality structure and texture priors learned by a well-trained deep generative model to enhance the realness and fidelity of the enhanced image.
Starting from this perspective, we design a zero-shot LIE model based on the Retinex theory,  which rapidly decomposes an image into a reflectance map and an illumination map. Instead of fine-tuning the parameters of the generative model, we backpropagate the gradients calculated from the loss functions to the input seeds, thus intactly retaining the deep generative knowledge learned from large-scale training samples. Our method iteratively optimizes on only one low-light image, then eliminating the need for low-light datasets at all.
In summary, the contributions of this work are listed as follows:

\begin{itemize}
    \item We propose a new perspective for LIE tasks, $i.e.$, strengthening Retinex decomposition with a well-trained generator, which uniquely leverages pre-trained knowledge to effectively address the challenges of insufficient information and severe feature degradation in single-image low-light conditions.
    \item We design a zero-shot LIE framework by embedding seed-optimization into a Retinex-based enhancement framework, which achieves fast convergence speed and great visual results. To the best of our knowledge, this is the first learning-based LIE approach that does not need to optimize the parameters of the enhancement model.
    \item We demonstrate the superiority of the proposed method through extensive experiments. With only limited time ($\textless$10s) for enhancement, the proposed method can even achieve superior performance against state-of-the-art methods trained on large-scale LIE datasets.
\end{itemize}

%% file: sec/2_related_works.tex
\section{Related Work}

\subsection{Dataset-based Low-light Enhancement}
$\indent$ In the research field of Low-Light Image Enhancement (LIE), a series of learning-based methods such as ~\cite{Supervised_2018Retinexnet,Supervised_zhang2019kind,Supervised_zhang2021kindpp,Supervised_cai2023Retinexformer,UnSupervised_jiang2021enlightengan,UnSupervised_ni2020uegan,UnSupervised_yang2023nerco,UnSupervised_liang2023iterative}, have achieved impressive results. 
To establish the mapping from low-light image to normal-light image, most of these methods involve supervised learning strategy, thus posing the serious reliance on the datasets containing numerous low-light and normal-light samples.
As an early supervised learning endeavor, RetinexNet~\cite{Supervised_2018Retinexnet} explores end-to-end Retinex decomposition. 
Advanced supervised methods such as KinD~\cite{Supervised_zhang2019kind}, Retinexformer~\cite{Supervised_cai2023Retinexformer} utilized convolutional neural network (CNN) and transformers to further explore illumination adjustment and reflectance restoration. However, the acquisition cost of paired images required by these methods is extremely high in real-world scenarios. 

In response to the challenge of acquiring paired training images, Unsupervised LIE methods have gained increasing attention. EnLightenGAN~\cite{UnSupervised_jiang2021enlightengan}, UEGAN~\cite{UnSupervised_ni2020uegan} employ adversarial training strategy, effectively utilizing unpaired data as both positive and negative samples with the assistance of a discriminator. Taking one step further, NeRCo~\cite{UnSupervised_yang2023nerco}, CLIP-LIT~\cite{UnSupervised_liang2023iterative} introduces multi-modal adversarial learning to low-light image enhancement, resulting in perceptually better results. 
To achieve satisfactory performance, these methods need large-scale training samples highly correlated with application scenarios in terms of both image content and lighting conditions, which is still costly in practice. 

\subsection{Zero-shot Low-light Enhancement}
$\indent$ To reduce the requirement for low-light datasets, many efforts~\cite{UnSupervised_guo2020zerodce,UnSupervised_li2021zerodcepp,Zeroshot_liang2022dunp,Zeroshot_fei2023gdp} have been directed toward zero-shot learning in recent years. There are two settings in zero-shot LIE: Firstly, methods such as Zero-DCE~\cite{UnSupervised_guo2020zerodce}, Zero-DCE++~\cite{UnSupervised_li2021zerodcepp} and RUAS~\cite{UnSupervised_liu2021ruas}, propose learning parameterized curves and architecture search for light enhancement without reliance on paired or unpaired data. However, these approaches still necessitate datasets with varied illumination for Optimization, which makes their performance still limited by the dataset distribution. Subsequently, other methods use only one low-light image for enhancement. Inspired by the concept of leveraging model structure prior, known as Deep Image Prior~\cite{ulyanov2018dip}, approaches like DUNP~\cite{Zeroshot_liang2022dunp} aim to thoroughly obviate the need for training data. These methods decompose the illumination and reflection components from a single image and subsequently enhance it based on the enhancement results. Nevertheless, DUNP employs randomly initialized weights, which results in a diminished correlation with real-world variance. Additionally, GDP~\cite{Zeroshot_fei2023gdp} leverages a degradation model and pre-trained Denoising Diffusion Probabilistic Model (DDPM)~\cite{ho2020ddpm} to directly obtain enhanced images. Although GDP attempts to introduce pre-trained models to mitigate optimization difficulties, its performance still heavily relies on the manually defined degradation model and extensive sampling time. In summary, while existing zero-shot learning methods alleviate the dependency on training data, their computational efficiency and perception of natural image characteristics remain unsatisfactory.

\subsection{Generative Models}
$\indent$ In recent years, generative models, including Generative Adversarial Networks (GANs) ~\cite{goodfellow2014gan,radford2015dcgan,karras2019stylegan,brock2018biggan} and Variational Autoencoders (VAEs) ~\cite{kingma2013vae,van2017vqvae,razavi2019vqvae2}, have achieved remarkable advancements within pioneering research trajectories in image synthesis. Their capability to produce high-fidelity generative outcomes has led to their widespread adoption across diverse domains. Notably, the latent diffusion approach ~\cite{rombach2022latentdiff} leverages the distinctive feature compression and quantization abilities of VQ-VAEs, thus enabling diffusion models to rapidly generate high-quality images. Within the domain of low-level vision tasks, codeformer~\cite{zhou2022codeformer} employs a codebook to rectify image features for face restoration, while RIDCP~\cite{wu2023ridcp} adopts a pre-trained vq-vae to improve image dehazing, effectively bridging the gap between synthetic and real-world data. Furthermore, GLEAN~\cite{chan2021glean} utilizes pre-trained StyleGAN to provide rich and diverse priors for image super-resolution. Motivated by the exciting performance of these approaches, we propose a method to efficiently utilize generative pre-trained knowledge to achieve zero-shot low-light image enhancement, setting a novel perspective in the utilization of generative knowledge without the need for additional fine-tuning or dataset-specific training.

%% file: sec/4_proposed_method.tex
\section{Method}
\subsection{Motivation}
\label{ssec:movitation}
\begin{figure}[!t]
\centering
\vspace{-0.3cm}
    \includegraphics[width=\linewidth]{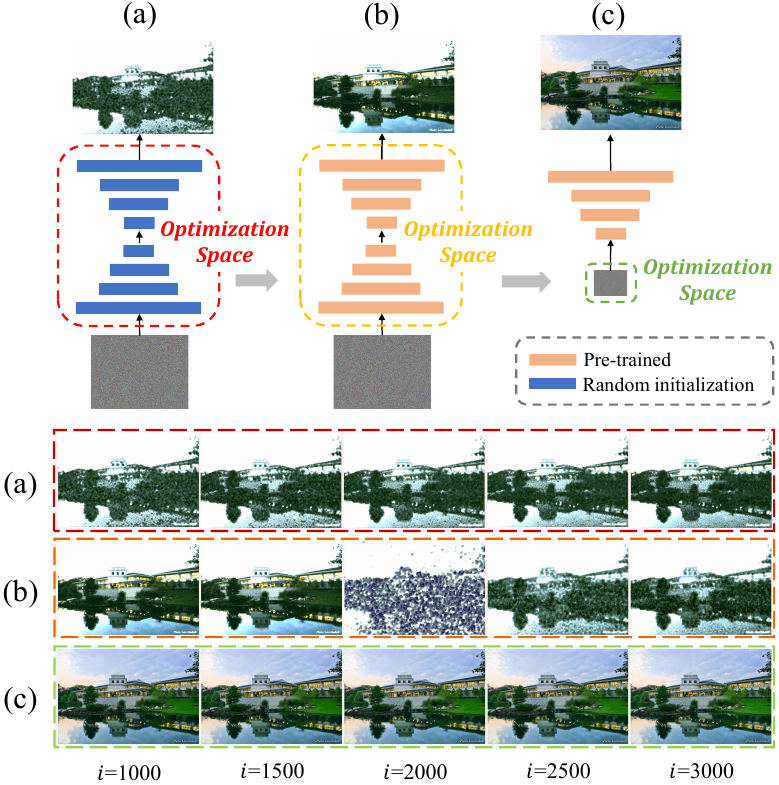}
    \caption{Retinex decomposition results in different optimization settings. The upper part of the Figure shows the mode of generating reflectance and the lower part shows the results of iterative optimization. The seed-optimization strategy shows better performance in terms of quality. More details will be shown in the supplementary material.}
    \label{fig:movitation}
\end{figure}

$\indent$ Generative models have achieved remarkable success in computer vision tasks. These models, adept at encapsulating the intricate distributions of natural image data due to their extensive training on vast image datasets, show promising versatility for various applications~\cite{rombach2022latentdiff,zhou2022codeformer}.

To mine the knowledge about natural images contained in the pre-trained generator and be able to assist in recovering low-light image features, we tried various ways of embedding them in low-light enhancement tasks. An intuitive way to leverage this generative knowledge is by taking a well-trained generator as the fundamental architecture and then fine-tuning it for image enhancement. However, such fine-tuning can disrupt the intricate generative knowledge cultivated on datasets of high-quality images. Our experiments, as illustrated in \Cref{fig:movitation} (a) and (b), reveal that although initializing with pre-trained weights enhances the decomposition quality in terms of detail and color, it falls short of mirroring the performance attainable in image generation domains.

Recognizing the limitations of conventional fine-tuning, we shift our strategy towards optimizing the input to the model rather than its parameters. As demonstrated in Figure \ref{fig:movitation} (c), our proposed methodology, which we refer to as the seed-optimization strategy, significantly outperforms standard fine-tuning techniques in image enhancement tasks. Motivated by this, we introduce a novel Retinex-based Low-light Image Enhancement (LIE) model grounded in the double-DIP framework. Contrary to traditional backpropagation methods that adjust model parameters, our approach optimizes the input seeds, thereby preserving the integrity of the generative knowledge acquired from well-lit image scenarios. This seed-centric optimization, guided by generative knowledge, facilitates the fast and superior enhancement of low-light images.

\begin{figure*}[!t]
\centering
\includegraphics[width=\textwidth]{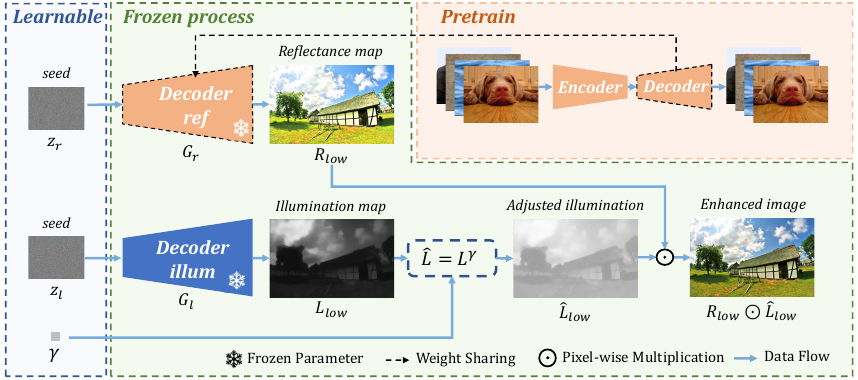}
\caption{Overview of Our Proposed Low-Light Enhancement Method. In the preparation phase, we utilize a conventionally trained image generator as the reflectance decoder, while the illumination decoder is initialized randomly. For an input low-light image, our design involves three optimizable seeds: $z_r$, $z_l$, and $\gamma$. $z_r$ and $z_l$ are dedicated to generating a reflectance map rich in detail and a comparatively smoother illumination map, respectively. Concurrently, $\gamma$ is employed for gamma correction of the image. The final enhancement result of the low-light image is achieved solely through iterative optimization of these three inputs.}
\label{fig:my_label}
\end{figure*}

\subsection{Retinex Model with Pre-trained Knowledge}

$\indent$ Given a low-light image, $I_{\text{low}}$, we apply Retinex theory to decompose it into a reflectance map, $R_{\text{low}}$, and an illumination map, $L_{\text{low}}$. This decomposition can be formally expressed as:
\begin{equation}
I_{\text{low}} = R_{\text{low}} \odot L_{\text{low}},
\label{eq:label1}
\end{equation}
where $\odot$ represents element-wise multiplication and $R_{\text{low}}$ denotes the reflectance map and $L_{\text{low}}$ denotes the illumination map. Correspondingly, the ideal normal-light image, $I_{\text{normal}}$, can be represented as:
\begin{equation}
I_{\text{normal}} = R_{\text{normal}} \odot L_{\text{normal}}.
\label{eq:label1}
\end{equation}
Based on the framework above, we use two generators, donated as $G_r$ and $G_l$, to generate the components $L$ and $R$ respectively, which can be formulated as:
\begin{equation}
I_{\text{low}} = G_r(z_r,\theta_r) \odot G_l(z_l,\theta_l),
\label{eq:label1}
\end{equation}
where $z_r$ and $z_l$ donate the random input seed, while $\theta_r$ and $\theta_l$ donate the weights of model.
The reflectance map contains abundant structure and texture details of the image, which follows the prior learned by the generative model. We employ a generative model pre-trained with weights $\tilde{\theta_r}$, which aims to effectively leverage generative knowledge from high-quality images to reconstruct the reflectance under varied lighting conditions. In contrast, the illumination map, primarily representing ambient light, deviates significantly from these high-quality image priors due to its inherent low-light characteristics. Therefore, we utilize a generator initialized with random weights, denoted as $\hat{\theta_l}$, to generate the illumination map. Now the decomposition process is defined as:
\begin{equation}
I_{\text{low}} = G_r(z_r,\tilde{\theta_r}) \odot G_l(z_l,\hat{\theta_l}),
\label{eq:label1}
\end{equation}
Having obtained the reflectance and illumination maps, we further apply a gamma transformation to effectively control the brightness of the enhanced result, bringing $L_{\text{low}}$ closer to $L_{\text{normal}}$. Unlike previous works that use specialized modules for adjusting gamma values, we integrate the gamma transformation as a learnable parameter within our enhancement framework. Denoting the gamma transformation factor as $\gamma$.The result of gamma transformation is donated as $\hat{L}_{\text{low}}$, and the final enhanced result is defined as:
\begin{equation}
I_{\text{result}} = G_r(z_r,\tilde{\theta_r}) \odot G_l(z_l,\hat{\theta_l})^\gamma.
\label{eq:label_result}
\end{equation}
This approach effectively incorporates generative knowledge into the Retinex model. Our next task is to define the optimization process for it.

\input{sec/tables/main_table}

\subsection{Prior-Constrained Seed Optimization}
$\indent$ According to equation \ref{eq:label_result}, for convenience, we simplify the expression of our model as:
\begin{equation}
I_{normal} = G(z, \theta),
\label{eq:label1}
\end{equation}
where $G$,$z$, and $\theta$ denote all the generators, random seeds, and parameters of the generator respectively. The initial values of random seeds are independently sampled from the same Gaussian distribution. For an $M \times N$ low-light image,$z$ is of size $M/{2^n} \times N/{2^n}$, where $n$ represents the number of upsampling layers in the decoder. 
The objective of our model is to minimize the discrepancy between the generated image and the target normal-light image using the loss function. To completely retain the deep generative priors learned from large-scale training samples, instead of fine-tuning the parameters of the generative model, we turn to optimize the inputs, which is represented as:
\begin{equation}
\minimize_{z} \mathcal{L}_{all}\left( G(z, \theta), I_{low} \right).
\label{eq:label2}
\end{equation}
In particular, we backpropagate the gradients calculated from the loss functions to the input seeds as:
\begin{equation}
z_{\text{new}} = z - lr \times \nabla z \mathcal{L}_{all}\left( G(z, \theta), I_{low} \right),
\label{eq:label3}
\end{equation}
where $lr$ is the learning rate and $\nabla z$ is the Derivative function.Based on this seed optimization strategy, we can leverage the generative knowledge to optimize the random seeds.

\subsection{Loss Functions}
$\indent$ To optimize the seeds, We use a series of concise and classic regularization terms to optimize the model. The overall loss function can be defined as:
\begin{equation}
\mathcal{L}_{all} = \lambda_{RE} \mathcal{L}_{RE} + \lambda_E \mathcal{L}_E + \lambda_{S} \mathcal{L}_{S} + \lambda_I \mathcal{L}_I,
\label{eq:label4}
\end{equation}
Among them, $\mathcal{L}_{RE}$ and $\mathcal{L}_E$ are used to perform standard Retinex decomposition, $\mathcal{L}_S$ is used to regularize the highly ill-posed Retinex framework, and $\mathcal{L}_I$ is used to enhance low-light images to a sufficient brightness, while $\lambda_{RE}$,$\lambda_E$,$\lambda_S$ and $\lambda_I$ is the balance factor.

\textbf{Reconstruction Loss.} To achieve accurate Retinex decomposition, the reconstruction loss is defined as:
\begin{equation}
\mathcal{L}_{RE} = \left\| I_{low} - G_r(z_r,\tilde{\theta_r}) \odot G_l(z_l,\hat{\theta_l}) \right\|^2.
\label{eq:label_rE}
\end{equation}
The reconstruction loss helps to achieve the standard Retinex decomposition.

\textbf{Illumination-consistency Loss.} To constrain the general structure of the illumination map, we define the illumination regularization loss as:
\begin{equation}
\mathcal{L}_E = \left\| \maximize_{c \in \{R,G,B\}} gauss(I_{low})^c - G_l(z_l,\hat{\theta_l}) \right\|,
\label{eq:label_E}
\end{equation}
where $gauss$ donates Gaussian blur with kernel size=25 and $\sigma=2.0$ and $c$ donates the channel of the input image.

\textbf{Smoothness Loss.} To regularize the highly ill-posed Retinex framework, we introduce the smoothness loss, which is defined as:
\begin{equation}
\mathcal{L}_S = \left\| \frac{\nabla G_l(z_l)}{\exp\left| \nabla G_R(z_r) \right|} \right\| + \tau \left\| \nabla G_R(z_r) \right\|,
\label{eq:label_s}
\end{equation}
where $\tau$ donates the balance factor. A weight matrix derived from the gradient map of the reflection map is used to smooth the illumination map over texture details while still maintaining overall structural boundaries.

\textbf{Illumination Control Loss.} The illumination loss encourages the overall illumination of the enhanced image closer to the desired illumination. To this end, we define the illumination control loss as:
\begin{equation}
\mathcal{L}_I = \left\| E - G_r(z_r,\tilde{\theta_r}) \odot G_l(z_l,\hat{\theta_l})^\gamma \right\|,
\label{eq:label_I}
\end{equation}
where $E$ is the well-exposedness level and is set to 0.6 in our experiments, following the setting in~\cite{UnSupervised_guo2020zerodce,UnSupervised_li2021zerodcepp}.

%% file: sec/tables/main_table.tex
\begin{table*}[]
\centering
\small
\renewcommand\arraystretch{1.2}
\resizebox{\textwidth}{!}{
\begin{tabular}{cccccccccccccccc}
\toprule
                                      &                                       & \multicolumn{4}{c}{\textbf{LOL}}                                                                                                                                                                                                    & \multicolumn{2}{c}{\textbf{MEF}}                                                                                    & \multicolumn{2}{c}{\textbf{LIME}}                                                                                            & \multicolumn{2}{c}{\textbf{DICM}}                                                                                            & \multicolumn{2}{c}{\textbf{NPE}}                                                                                    &                                                              &                                                           \\ \cline{3-14}
                                      & \multirow{-2}{*}{\textbf{Method}}     & \textbf{P↑}                                                   & \textbf{S↑}                                                  & \textbf{N↓}                                                  & \textbf{M↑}                           & \textbf{N↓}                                         & \textbf{M↑}                                                   & \textbf{N↓}                                                  & \textbf{M↑}                                                   & \textbf{N↓}                                                  & \textbf{M↑}                                                   & \textbf{N↓}                                         & \textbf{M↑}                                                   & \multirow{-2}{*}{\textbf{Rank}}                              & \multirow{-2}{*}{\textbf{RoR}}                            \\ \hline
                                      & Retinex-Net                           & 16.77                                                         & 0.42                                                         & 8.87                                                         & 57.26                                 & 4.40                                                & 66.71                                                         & 4.60                                                         & 62.90                                                         & 4.43                                                         & 63.11                                                         & 4.47                                                & 65.16                                                         & 8.42                                                         & 10                                                        \\
                                      & KinD++                                & 21.80                                                         & 0.83                                                         & 5.12                                                         & 70.54                                 & 3.76                                                & 66.11                                                         & 4.73                                                         & 64.00                                                         & 3.78                                                         & 62.81                                                         & 3.76                                                & 65.53                                                         & {\color[HTML]{3166FF} 5.00}                                  & {\color[HTML]{3166FF} 2}                                  \\
\multirow{-3}{*}{Supervised Training} & SNR                                   & 24.61                                                         & 0.84                                                         & 5.17                                                         & 59.79                                 & 4.15                                                & 57.49                                                         & 5.69                                                         & 52.52                                                         & 4.64                                                         & 50.19                                                         & 5.70                                                & 50.67                                                         & 10.08                                                        & 12                                                        \\ \hline
                                      & EnlightenGan                          & 17.48                                                         & 0.65                                                         & 4.68                                                         & 56.60                                 & 3.20                                                & 63.03                                                         & 3.66                                                         & 59.07                                                         & 3.56                                                         & 57.03                                                         & 3.70                                                & 59.41                                                         & 6.67                                                         & 4                                                         \\
                                      & SCL-LLE                               & 12.42                                                         & 0.52                                                         & 7.63                                                         & 56.39                                 & 3.28                                                & 65.01                                                         & 3.77                                                         & 61.45                                                         & 3.57                                                         & 60.43                                                         & 3.59                                                & 63.75                                                         & 6.92                                                         & 6                                                         \\
\multirow{-3}{*}{Unpaired Training}   & NeRCo                                 & 19.74                                                         & 0.80                                                         & 3.40                                                         & 69.87                                 & 3.76                                                & 61.91                                                         & 3.59                                                         & 61.24                                                         & 3.80                                                         & 64.08                                                         & 3.69                                                & 65.55                                                         & 5.33                                                         & 3                                                         \\ \hline
                                      & RUAS                                  & {\color[HTML]{3166FF} 16.40}                                  & 0.50                                                         & 6.34                                                         & 59.34                                 & 5.47                                                & 55.53                                                         & 5.37                                                         & 55.43                                                         & 7.20                                                         & 50.41                                                         & 7.16                                                & 50.78                                                         & 11.75                                                        & 14                                                        \\
                                      & SCI                                   & 14.78                                                         & 0.52                                                         & 7.87                                                         & 57.12                                 & 3.62                                                & 62.59                                                         & 4.18                                                         & 59.11                                                         & 4.13                                                         & 53.90                                                         & 4.02                                                & 56.94                                                         & 10.25                                                        & 13                                                        \\
                                      & SGZ                                   & 15.93                                                         & {\color[HTML]{3166FF} 0.57}                                  & 7.82                                                         & 56.53                                 & 3.34                                                & 63.87                                                         & 3.91                                                         & 60.41                                                         & 3.56                                                         & 57.42                                                         & {\color[HTML]{FE0000} \textbf{3.52}}                & 60.82                                                         & 7.25                                                         & 8                                                         \\
                                      & ZeroDCE++                             & 14.86                                                         & 0.56                                                         & 7.77                                                         & 55.85                                 & 3.32                                                & 65.13                                                         & 3.79                                                         & 60.73                                                         & 3.57                                                         & 59.31                                                         & 3.59                                                & 63.13                                                         & 6.67                                                         & 4                                                         \\
                                      & GDP                                   & 15.89                                                         & 0.54                                                         & 6.13                                                         & {\color[HTML]{3166FF} 60.25}          & 4.08                                                & 59.64                                                         & 4.40                                                         & 59.69                                                         & 4.11                                                         & 58.83                                                         & 3.67                                                & 59.91                                                         & 8.67                                                         & 11                                                        \\
                                      & DUNP                                  & 13.21                                                         & 0.46                                                         & {\color[HTML]{3166FF} 4.27}                                  & {\color[HTML]{FE0000} \textbf{67.06}} & {\color[HTML]{FE0000} \textbf{3.10}}                & 65.08                                                         & 4.12                                                         & 56.90                                                         & 3.65                                                         & 56.37                                                         & 3.81                                                & 68.03                                                         & 7.08                                                         & 7                                                         \\
\multirow{-7}{*}{Zero-Shot}           & \cellcolor[HTML]{EFEFEF}\textbf{Ours} & \cellcolor[HTML]{EFEFEF}{\color[HTML]{FE0000} \textbf{18.10}} & \cellcolor[HTML]{EFEFEF}{\color[HTML]{FE0000} \textbf{0.75}} & \cellcolor[HTML]{EFEFEF}{\color[HTML]{FE0000} \textbf{2.80}} & \cellcolor[HTML]{EFEFEF}58.98         & \cellcolor[HTML]{EFEFEF}{\color[HTML]{3166FF} 3.16} & \cellcolor[HTML]{EFEFEF}{\color[HTML]{FE0000} \textbf{65.19}} & \cellcolor[HTML]{EFEFEF}{\color[HTML]{FE0000} \textbf{3.64}} & \cellcolor[HTML]{EFEFEF}{\color[HTML]{FE0000} \textbf{63.12}} & \cellcolor[HTML]{EFEFEF}{\color[HTML]{FE0000} \textbf{3.43}} & \cellcolor[HTML]{EFEFEF}{\color[HTML]{FE0000} \textbf{65.19}} & \cellcolor[HTML]{EFEFEF}{\color[HTML]{3166FF} 3.65} & \cellcolor[HTML]{EFEFEF}{\color[HTML]{FE0000} \textbf{68.24}} & \cellcolor[HTML]{EFEFEF}{\color[HTML]{FE0000} \textbf{2.92}} & \cellcolor[HTML]{EFEFEF}{\color[HTML]{FE0000} \textbf{1}}

\\
\bottomrule
\end{tabular}
}
\caption{\underline{P}SNR/\underline{S}SIM/\underline{N}IQE/\underline{M}USIQ scores on five datasets. The best and second-best Zero-Shot methods are highlighted in red and blue respectively. The overall 'Rank' is calculated by averaging the per-dataset average rankings among every method, while 'RoR' denotes the rank of the overall Rank, the best and second-best methods are also highlighted in red and blue. ↑ (↓) means higher (lower) is better.}
\label{tab:main_table}
\vspace{-0.3cm}
\end{table*}

%% file: sec/5_experiments.tex
\section{Experiments}
\begin{figure*}[!t]
    \centering
    \vspace{-0.3cm}
    \includegraphics[width=\linewidth]{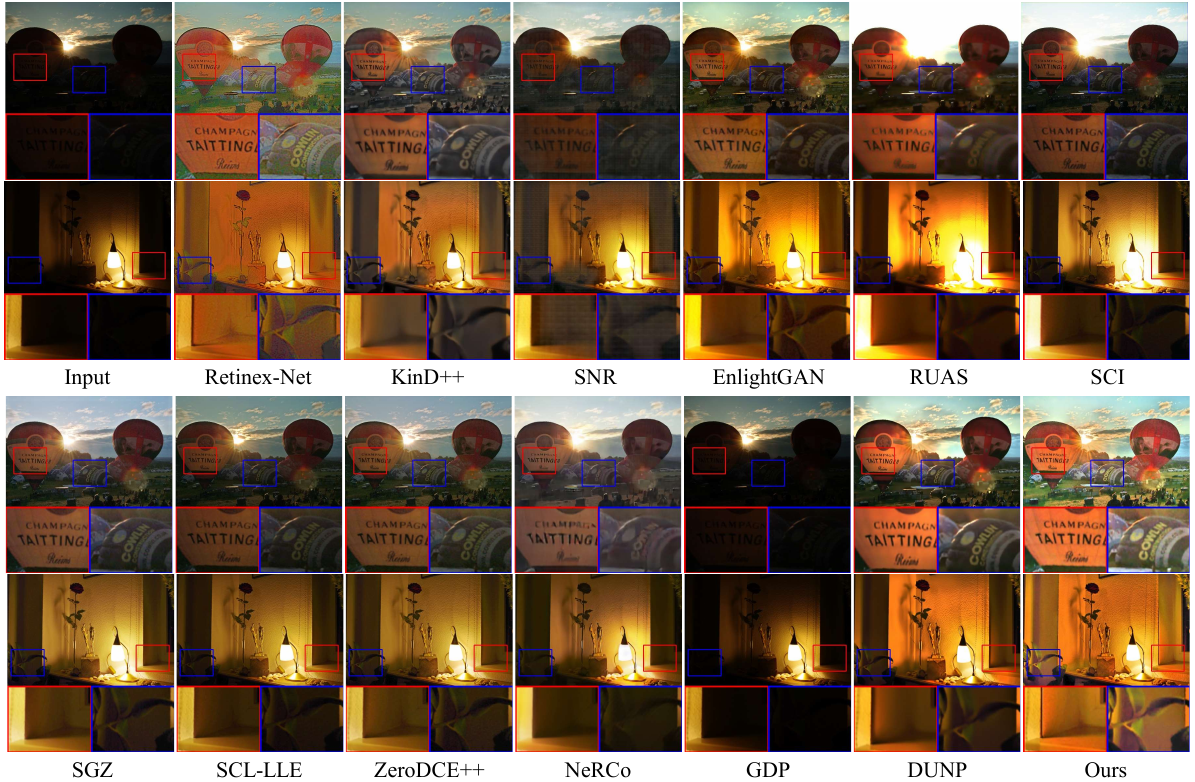}
    \caption{Visual quality comparison with several low-light image enhancement state-of-the-art methods. Our approach demonstrates strong detail and color recovery capabilities.}
    \label{fig:compare}
    \vspace{-0.3cm}
\end{figure*}

\subsection{Implementation Details}

$\indent$ The proposed method is based on pytorch and runs with an NVIDIA TITAN RTX GPU and an Intel(R) Xeon(R) Gold 6252 CPU @ 2.10GHz. The Adam optimizer is set to do the train with learning rate of 1e-2. For the non-referenced datasets, each image undergoes an optimization process over 5000 epochs, and $\tau$ is set to 0.2, while images from the LOL dataset are optimized for 2500 epochs, and $\tau$ is set to 0.6. The learnable param gamma is initialized as 0.5, and $\lambda_{RE}$, $\lambda_E$, $\lambda_S$, and $\lambda_I$ is set to be 12, 0.05, 0.03, and 0.01 respectively.

For the pre-trained model, we use the VQ-VAE-2 model with the Adam optimizer at learning rate of 3e-4 and train on the FFHQ~\cite{karras2019stylegan} and ImageNet~\cite{deng2009imagenet} dataset with batchsize of 128 for 600 epochs. In practice, we find that different types of generative models can achieve good results under their respective recommended data pipelines, which will be demonstrated in Sec~\ref{sec:abl} and supplementary material.

\subsection{Datasets}
$\indent$ We choose 5 referenced or non-reference datasets to comprehensively evaluate various methods in low-light enhancement tasks: a) LOL~\cite{Supervised_2018Retinexnet} included 15 indoor images in severely underexposed conditions. b) NPE~\cite{wang2013npe} contains 8 low-light nature images. c) MEF~\cite{ma2015mef17} comprised 17 low-light images in different scenes. d)LIME~\cite{guo2016lime} featured 10 dark scene images of different resolutions. e) DICM~\cite{lee2013dicm} contains 69 images from low-light and normal-light conditions. We use the most universal NIQE~\cite{mittal2012niqe} and the state-of-the-art MUSIQ~\cite{ke2021musiq} metrics to compare the performance of all methods on these datasets. For LOL datasets with paired images, we additionally use PSNR and SSIM~\cite{wang2004ssim} to evaluate the method based on reference images.

\subsection{Comparison with State-of-the-Art Methods}

$\indent$ For a more comprehensive analysis, we compare our method with three advanced supervised learning methods (i.e., RetinexNet ~\cite{Supervised_2018Retinexnet}, KinD++ ~\cite{Supervised_zhang2021kindpp}, and SNR ~\cite{Supervised_xu2022snr}), and nine unsupervised learning methods, including DUNP ~\cite{Zeroshot_liang2022dunp}, GDP ~\cite{Zeroshot_fei2023gdp}, Zero-DCE++ ~\cite{UnSupervised_guo2020zerodce}, SCL-LLE ~\cite{UnSupervised_liang2022scllle}, RUAS ~\cite{UnSupervised_liu2021ruas}, EnlightGAN ~\cite{UnSupervised_jiang2021enlightengan},  SGZ ~\cite{UnSupervised_zheng2022SGZ}, NeRCo~\cite{UnSupervised_yang2023nerco} and SCI ~\cite{UnSupervised_ma2022sci}.

\textbf{Quantitative Analysis.} We evaluated the performance of various methods using their official pre-trained models and publicly available code. As presented in Table \ref{tab:main_table}, achieves the highest average ranking across all benchmarks on full-reference and no-reference metrics. When compared to recent prominent same-category techniques like DUNP~\cite{Zeroshot_liang2022dunp} and GDP~\cite{Zeroshot_fei2023gdp}, our method demonstrates distinct advantages. We incorporate more robust generative priors than DUNP, leveraging additional pre-trained knowledge. Furthermore, thanks to the well-established Retinex model, our method consistently outperforms GDP, especially on a series of challenging datasets. For a more intuitive visual comparison, we provide results from all methods, including DUNP, and GDP, in scenarios characterized by significant brightness degradation in the subsequent sections. Compared with other zero-shot methods (where RUAS, Zero-DCE, SCI, and SGZ use low-light datasets), our method achieves SOTA on 75\% of the metrics, and even when compared with dataset-based methods achieved first overall ranking.

\textbf{Qualitative Analysis.} For a more visual understanding, we present the visual outcomes of all methodologies in Figure \ref{fig:compare}. It can be seen that recent deep learning-based techniques often struggle to produce enhanced images with natural illumination. Specifically, the outputs from EnlightGAN~\cite{UnSupervised_jiang2021enlightengan}, KinD++~\cite{UnSupervised_ni2020uegan}, and DUNP~\cite{Zeroshot_liang2022dunp} exhibit uneven illumination and poor details. There are noticeable over-exposures in the results from RUAS~\cite{UnSupervised_liu2021ruas} and SCI~\cite{UnSupervised_ma2022sci}. Both GDP~\cite{Zeroshot_fei2023gdp} and SNR~\cite{Supervised_xu2022snr} fall short in effectively enhancing extremely dark regions. RetinexNet~\cite{Supervised_2018Retinexnet}'s output is characterized by a pronounced animated style with non-authentic delineations. Similarly, Zero-DCE++~\cite{UnSupervised_guo2020zerodce} and others occasionally introduce unnatural noise and pronounced bright spots. By comparison, our model realizes the best visual quality with prominent contrast and vivid colors, while restoring intricate details. Extended results can be found in the supplementary material.

\begin{figure}[h]
    \centering
    \vspace{-0.3cm}
    \includegraphics[width=\linewidth]{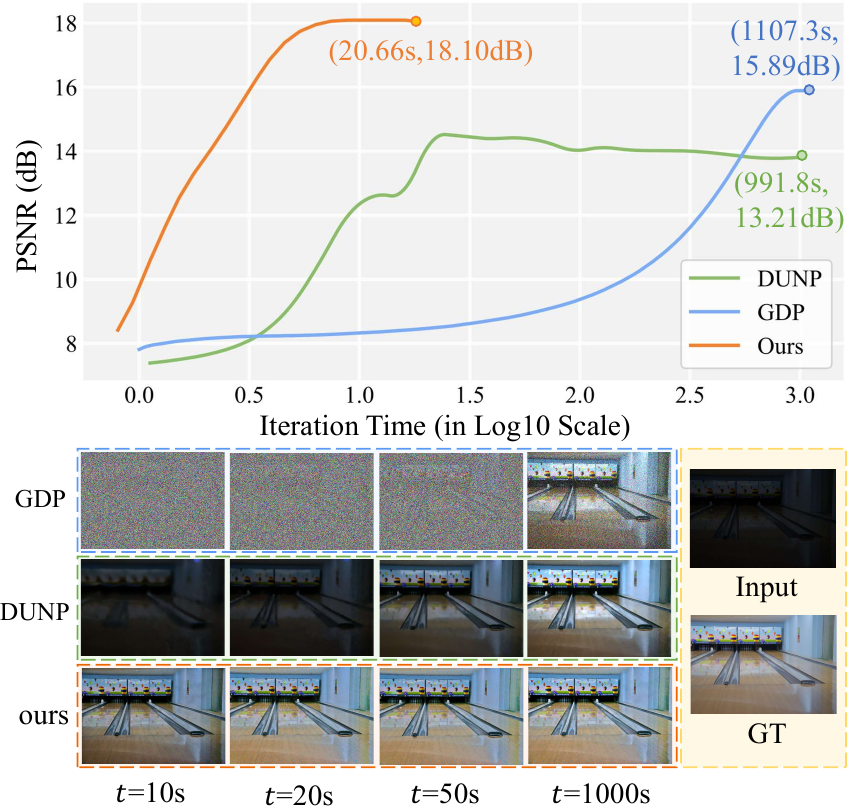}
    \caption{Comparison of efficiency with other methods that can be trained with a low-light image itself. All results are obtained on the LOL dataset using the same computing resources.}
    \label{fig:time}
\end{figure}

\textbf{Computational Efficiency.} In addressing the computational efficiency of zero-shot low-light enhancement methods, Figure~\ref{fig:time} provides a comparative analysis of the runtime for several state-of-the-art techniques that do not require training on low-light datasets. These methods were evaluated under uniform conditions at a resolution of 600$\times$400 pixels, following their respective recommended configurations. The evaluation encompasses the performance metrics and datasets previously detailed. Notably, as shown in Table \ref{tab:running time} our proposed method, even when limited to 7.3 seconds, outperforms others in terms of accuracy and computational efficiency. 
\input{sec/tables/time}

\subsection{Ablation Studies}
\label{sec:abl}
$\indent$ We conducted extensive ablation Studies to validate the efficacy of our proposed perspective and evaluate the contribution of each model component. All evaluations were carried out on the well-referenced LOL dataset.

\begin{figure}[!t]
\centering
\vspace{-0.3cm}
\includegraphics[width=\linewidth]{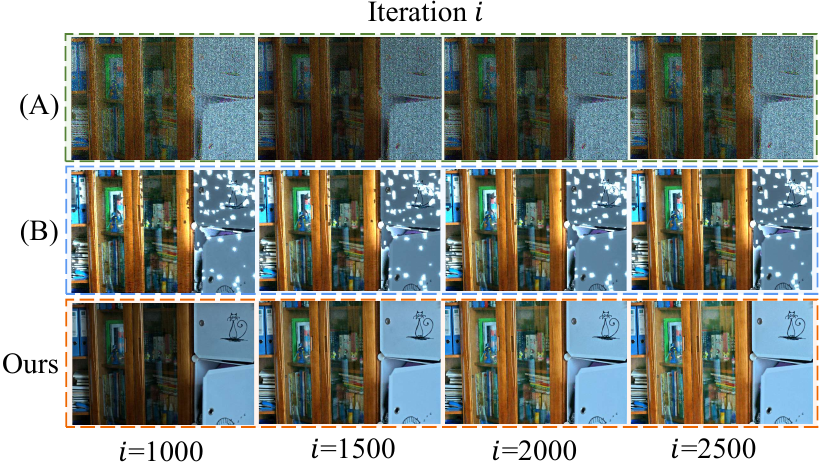}
    \caption{Ablation Study of Pre-training. (A): Random Initialization of the Reflectance Decoder. (B): Initialization of both the Illumination Decoder and Reflectance Decoder with Weights from Pre-trained Models.}
    \label{fig:pretrain}
\vspace{-0.3cm}
\end{figure}

\textbf{Effect of Pre-trained Weights.} We further delve into the impact of using pre-trained weights in our method by conducting experiments under three configurations: A): Random initialization for all decoders. B): Load pre-trained models for both reflectance decoder and illumination decoder. Ours: Load pre-trained models for reflectance decoder only. Figure \ref{fig:pretrain} graphically illustrates the outcomes for these settings. 
From the results in Figure, it is evident that our approach markedly outperforms Setting A. This showcases the strength of pre-trained priors derives from large-scale generative models. In contrast to Setting A, our method achieves a modest improvement in performance. This can be attributed to the discrepancy between the single-channel and smooth characteristics of the illumination map and the pre-trained weights, which leads to the production of inferior-quality illumination maps. The quantitative results in Table \ref{tab:abl_pretrain} further indicate the consequential role that pre-trained weights play in the enhancement model's performance.
\input{sec/tables/abl_pretrain}

\begin{figure}[!t]
\centering
\vspace{-0.0cm}
    \includegraphics[width=\linewidth]{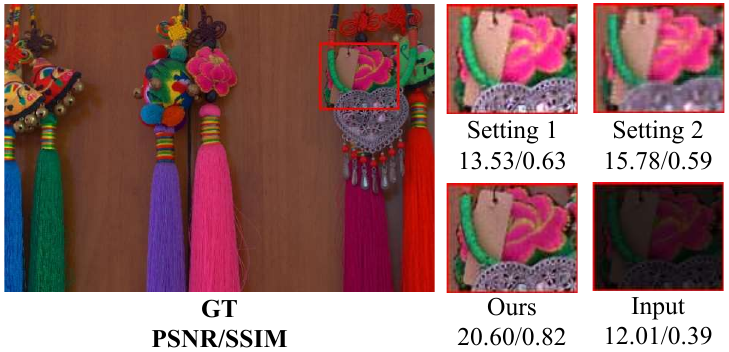}
    \caption{Ablation Study of Optimization Mode. (1): Consistent Seed with Model Fine-tuning. (2): Joint Optimization of Seed and Model Parameters.}
    \label{fig:optimode}
    \vspace{-0.3cm}
\end{figure}

\textbf{Optimization Mode.} To evaluate the benefits of retaining the deep generative principles learn from normal light images, we conduct experiments with four settings: (1) optimizing only the gamma, reflectance decoder, and illumination decoder; (2) optimizing all the inputs, and decoders; (3) using the recommended optimization settings from our proposed method. Through experiments, as shown in Table~\ref{tab:abl_optimode} and Figure~\ref{fig:optimode}, it can be seen that Setting 1 is similar to traditional DIP methods in three metrics, and directly using pre-trained models does not significantly improve model performance; Compare to Setting 1, Setting 2 additionally optimizes the input, but this training method actually makes the coordination between the model and input noise more chaotic, leading to further performance degradation; Our approach directly optimizes the input and achieves the best results, which is consistent with our observation in sec \ref{ssec:movitation}.
\input{sec/tables/abl_optimode}

\textbf{Pre-trained Model Selection.} To elucidate the benefits of our proposed training framework, we train our enhancement network utilizing various generative pre-trained models from both GAN~\cite{ying2022eagan,radford2015dcgan} and VAE~\cite{kingma2013vae,razavi2019vqvae2} families. Remarkably, as shown in Table~\ref{tab:abl_modelselect}, VQ-VAE-2 pre-trained model outperforms the others, albeit other models demonstrate near-par performance. This suggests that our approach is versatile enough to be compatible with diverse image generation techniques, even those not explicitly designed for low-light enhancement tasks.
\input{sec/tables/abl_modelselect}

\textbf{Analysis of Loss Contributions.} We present the results of our method trained without different losses in Table \ref{tab:abl_loss}. The result shows that removing $\mathcal{L}_S$ and $\mathcal{L}_E$ will reduce the quality of the generated image. Visual comparison diagrams are further given in the supplementary material.
\input{sec/tables/abl_loss}

%% file: sec/tables/time.tex
\begin{table}[h]
\centering
\setlength{\abovecaptionskip}{0.1cm} 
\setlength{\belowcaptionskip}{0.1cm} 
\setlength{\tabcolsep}{5pt} 
\resizebox{0.9\columnwidth}{!}{
\begin{tabular}{lcccc}
\toprule
\textbf{Method}   & \textbf{Total Time} & \textbf{Iteration Time}           & \textbf{Rank} & \textbf{RoR} \\ \hline
GDP               & 1107.2            & 1.10702                            & 10.17         & 9                   \\
DUNP              & 991.9            & 0.06611                           & 9.08          & 6           \\ 
Ours(900 iters)   & \textbf{7.3}     & \textbf{0.00813}                  & 3.95          & \textbf{1}   \\
Ours(2500 iters)                 & 20.3                                 & \textbf{0.00813}        & 3.42                           & \textbf{1}   \\
Ours(5000 iters)                 & 40.6                                 & \textbf{0.00813}                         & 3.39                           & \textbf{1}                    \\
Ours(10000 iters) & 81.3             & \textbf{0.00813}                 & \textbf{3.37} & \textbf{1}   \\ \hline
\end{tabular}}
\caption{Comparison of Processing Time: The average time(GPU seconds) for enhancing one low-light image with the size of 600 $\times$ 400. 'Rank' and 'RoR' follow the definition in Table ~\ref{tab:main_table}. The best results are marked in bold.}
\label{tab:running time}
\vspace{-0.3cm}
\end{table}

%% file: sec/tables/abl_pretrain.tex
\begin{table}[H]
\centering
\setlength{\abovecaptionskip}{0.1cm} 
\setlength{\belowcaptionskip}{0.1cm} 
\setlength{\tabcolsep}{5pt} 
\resizebox{\columnwidth}{!}{
\begin{tabular}{ccccccc}
\toprule
\textbf{Setting} & $G_r$ & $G_l$ & \textbf{PSNR↑} & \textbf{SSIM↑} & \textbf{NIQE↓} & \textbf{MUSIQ↑} \\ \hline
\#A     &              &             & 14.44 & 0.46 & 3.09 & 57.60  \\
\#B     &  \checkmark  &  \checkmark & 16.83 & 0.65 & 3.35 & 58.42  \\
Ours    &  \checkmark  &             & \textbf{18.10} & \textbf{0.75} & \textbf{2.80} & \textbf{58.98}  \\ \hline
\end{tabular}
}
\caption{Quantitative Analysis from Ablation Studies Demonstrating the Impact of Pretrained Weights on reflectance and Illumination Decoder. $\checkmark$ means placing load pretrained weight to this generator. The best results are marked in bold.}
\label{tab:abl_pretrain}
\vspace{-0.3cm}
\end{table}

%% file: sec/tables/abl_optimode.tex
\begin{table}[h]
\centering
\setlength{\abovecaptionskip}{0.1cm} 
\setlength{\belowcaptionskip}{0.1cm} 
\renewcommand\arraystretch{1.1}
\setlength{\tabcolsep}{5pt} 
\resizebox{\columnwidth}{!}{
\begin{tabular}{lcccccc}
\toprule
\textbf{Setting} & $z$ & $\theta$ & \textbf{PSNR↑} & \textbf{SSIM↑} & \textbf{NIQE↓} & \textbf{MUSIQ↑} \\ \hline
\#1     &              & \checkmark   & 15.51 & 0.68 & 3.61 & 56.92  \\
\#2     & \checkmark   & \checkmark   & 14.88 & 0.61 & 4.08 & 56.58  \\
Ours    & \checkmark   &    & \textbf{18.10} & \textbf{0.75} & \textbf{2.80} & \textbf{58.98}  \\ \hline 
\end{tabular}
}
\caption{Quantitative Analysis from Ablation Studies Demonstrating the Impact of Optimization mode. $\checkmark$ means placing this parameter in the optimizer for iterative training. The best results are marked in bold.}
\label{tab:abl_optimode}
\vspace{-0.3cm}
\end{table}

%% file: sec/tables/abl_modelselect.tex
\begin{table}[h]
\centering
\setlength{\abovecaptionskip}{0.1cm} 
\setlength{\belowcaptionskip}{0.1cm} 
\setlength{\tabcolsep}{5pt} 
\resizebox{0.8\columnwidth}{!}{
\begin{tabular}{lcccc}
\toprule
         & \textbf{PSNR↑}    & \textbf{SSIM↑}   & \textbf{NIQE↓}   & \textbf{MUSIQ↑} \\ \hline
DCGAN    & 16.09           & 0.69          & 3.61          & 58.63          \\
EAGAN & 16.94           & 0.74          & 4.96            & 55.04          \\
VAE      & 17.19          & 0.72          & 4.19          & 58.22          \\
VQ-VAE-2  & \textbf{18.10} & \textbf{0.75} & \textbf{2.80} & \textbf{58.98}          \\ \hline
\end{tabular}
}
\caption{Selection of Pre-trained Models. We place different pre-trained models into the proposed framework and retest them on the LOL dataset for low-light enhancement. The best results are marked in bold.}
\label{tab:abl_modelselect}
\vspace{-0.3cm}
\end{table}

%% file: sec/tables/abl_loss.tex
\begin{table}[H]
\centering
\setlength{\abovecaptionskip}{0.1cm} 
\setlength{\belowcaptionskip}{0.1cm} 
\setlength{\tabcolsep}{5pt} 
\resizebox{0.8\columnwidth}{!}{
\begin{tabular}{cccccc}
\toprule
$\mathcal{L}_E$                & $\mathcal{L}_S$                & \textbf{PSNR↑} & \textbf{SSIM↑} & \textbf{NIQE↓} & \textbf{MUSIQ↑} \\ \hline
\multicolumn{1}{l}{} & \multicolumn{1}{l}{} & 10.39                           & 0.53                            & 3.78                            & 55.02                            \\
\checkmark                    &                      & 16.09                           & 0.69                            & 3.61                            & 56.35                            \\
                     & \checkmark                      & 14.56                           & 0.65                            & 2.99                            & 57.10                            \\
\checkmark                      & \checkmark                      & \textbf{18.10} & \textbf{0.75}  & \textbf{2.80}  & \textbf{58.98}  \\ \hline
\end{tabular}}
\caption{Quantitative Analysis from Ablation Studies Demonstrating the Impact of Losses. $\checkmark$ means using this loss. The best results are marked in bold.}
\label{tab:abl_loss}
\vspace{-0.3cm}
\end{table}

%% file: sec/6_conclusion.tex
\section{Conclusion}
$\indent$ In this work, we introduce an innovative zero-shot low-light enhancement approach that integrates a pre-trained generative model into a Retinex-based enhancement framework. This integration not only accelerates convergence but also significantly improves visual outcomes in low-light conditions. We initially observed that generative knowledge can be effectively applied to Retinex decomposition and reflection map reconstruction. Subsequently, we employ a seed optimization strategy to preserve the advantages of generative knowledge, leading to the development of an efficient Retinex decomposition framework. Our approach notably reduces the reliance on extensive low-light datasets, demonstrating its effectiveness across a variety of lighting conditions and scenes. Through comprehensive experiments in diverse scenarios, our method has shown remarkable generalization capabilities. It is poised to adapt to a range of existing and future generative models, underlining its potential for broad applicability in the field of low-light image enhancement. 

%% file: main.bbl
\begin{thebibliography}{45}
\providecommand{\natexlab}[1]{#1}
\providecommand{\url}[1]{\texttt{#1}}
\expandafter\ifx\csname urlstyle\endcsname\relax
  \providecommand{\doi}[1]{doi: #1}\else
  \providecommand{\doi}{doi: \begingroup \urlstyle{rm}\Url}\fi

\bibitem[Brock et~al.(2018)Brock, Donahue, and Simonyan]{brock2018biggan}
Andrew Brock, Jeff Donahue, and Karen Simonyan.
\newblock Large scale gan training for high fidelity natural image synthesis.
\newblock \emph{arXiv preprint arXiv:1809.11096}, 2018.

\bibitem[Cai et~al.(2023)Cai, Bian, Lin, Wang, Timofte, and Zhang]{Supervised_cai2023Retinexformer}
Yuanhao Cai, Hao Bian, Jing Lin, Haoqian Wang, Radu Timofte, and Yulun Zhang.
\newblock Retinexformer: One-stage retinex-based transformer for low-light image enhancement.
\newblock \emph{arXiv preprint arXiv:2303.06705}, 2023.

\bibitem[Chan et~al.(2021)Chan, Wang, Xu, Gu, and Loy]{chan2021glean}
Kelvin~CK Chan, Xintao Wang, Xiangyu Xu, Jinwei Gu, and Chen~Change Loy.
\newblock Glean: Generative latent bank for large-factor image super-resolution.
\newblock In \emph{Proceedings of the IEEE/CVF conference on computer vision and pattern recognition}, pages 14245--14254, 2021.

\bibitem[Deng et~al.(2009)Deng, Dong, Socher, Li, Li, and Fei-Fei]{deng2009imagenet}
Jia Deng, Wei Dong, Richard Socher, Li-Jia Li, Kai Li, and Li Fei-Fei.
\newblock Imagenet: A large-scale hierarchical image database.
\newblock In \emph{2009 IEEE conference on computer vision and pattern recognition}, pages 248--255. Ieee, 2009.

\bibitem[Fei et~al.(2023)Fei, Lyu, Pan, Zhang, Yang, Luo, Zhang, and Dai]{Zeroshot_fei2023gdp}
Ben Fei, Zhaoyang Lyu, Liang Pan, Junzhe Zhang, Weidong Yang, Tianyue Luo, Bo Zhang, and Bo Dai.
\newblock Generative diffusion prior for unified image restoration and enhancement.
\newblock In \emph{Proceedings of the IEEE/CVF Conference on Computer Vision and Pattern Recognition}, pages 9935--9946, 2023.

\bibitem[Fu et~al.(2023)Fu, Yang, Tu, Huang, Ding, and Ma]{UnSupervised_fu2023PairedInstance}
Zhenqi Fu, Yan Yang, Xiaotong Tu, Yue Huang, Xinghao Ding, and Kai-Kuang Ma.
\newblock Learning a simple low-light image enhancer from paired low-light instances.
\newblock In \emph{Proceedings of the IEEE/CVF Conference on Computer Vision and Pattern Recognition}, pages 22252--22261, 2023.

\bibitem[Goodfellow et~al.(2014)Goodfellow, Pouget-Abadie, Mirza, Xu, Warde-Farley, Ozair, Courville, and Bengio]{goodfellow2014gan}
Ian Goodfellow, Jean Pouget-Abadie, Mehdi Mirza, Bing Xu, David Warde-Farley, Sherjil Ozair, Aaron Courville, and Yoshua Bengio.
\newblock Generative adversarial nets.
\newblock \emph{Advances in neural information processing systems}, 27, 2014.

\bibitem[Guo et~al.(2020)Guo, Li, Guo, Loy, Hou, Kwong, and Cong]{UnSupervised_guo2020zerodce}
Chunle Guo, Chongyi Li, Jichang Guo, Chen~Change Loy, Junhui Hou, Sam Kwong, and Runmin Cong.
\newblock Zero-reference deep curve estimation for low-light image enhancement.
\newblock In \emph{Proceedings of the IEEE/CVF conference on computer vision and pattern recognition}, pages 1780--1789, 2020.

\bibitem[Guo et~al.(2016)Guo, Li, and Ling]{guo2016lime}
Xiaojie Guo, Yu Li, and Haibin Ling.
\newblock Lime: Low-light image enhancement via illumination map estimation.
\newblock \emph{IEEE Transactions on image processing}, 26\penalty0 (2):\penalty0 982--993, 2016.

\bibitem[Ho et~al.(2020)Ho, Jain, and Abbeel]{ho2020ddpm}
Jonathan Ho, Ajay Jain, and Pieter Abbeel.
\newblock Denoising diffusion probabilistic models.
\newblock \emph{Advances in neural information processing systems}, 33:\penalty0 6840--6851, 2020.

\bibitem[Islam et~al.(2020)Islam, Edge, Xiao, Luo, Mehtaz, Morse, Enan, and Sattar]{Degradation_islam2020semantic}
Md~Jahidul Islam, Chelsey Edge, Yuyang Xiao, Peigen Luo, Muntaqim Mehtaz, Christopher Morse, Sadman~Sakib Enan, and Junaed Sattar.
\newblock Semantic segmentation of underwater imagery: Dataset and benchmark.
\newblock In \emph{2020 IEEE/RSJ International Conference on Intelligent Robots and Systems (IROS)}, pages 1769--1776. IEEE, 2020.

\bibitem[Jiang et~al.(2023)Jiang, Luo, Han, Fan, and Liu]{Supervised_jiang2023wavelet}
Hai Jiang, Ao Luo, Songchen Han, Haoqiang Fan, and Shuaicheng Liu.
\newblock Low-light image enhancement with wavelet-based diffusion models.
\newblock \emph{arXiv preprint arXiv:2306.00306}, 2023.

\bibitem[Jiang et~al.(2021)Jiang, Gong, Liu, Cheng, Fang, Shen, Yang, Zhou, and Wang]{UnSupervised_jiang2021enlightengan}
Yifan Jiang, Xinyu Gong, Ding Liu, Yu Cheng, Chen Fang, Xiaohui Shen, Jianchao Yang, Pan Zhou, and Zhangyang Wang.
\newblock Enlightengan: Deep light enhancement without paired supervision.
\newblock \emph{IEEE transactions on image processing}, 30:\penalty0 2340--2349, 2021.

\bibitem[Karras et~al.(2019)Karras, Laine, and Aila]{karras2019stylegan}
Tero Karras, Samuli Laine, and Timo Aila.
\newblock A style-based generator architecture for generative adversarial networks.
\newblock In \emph{Proceedings of the IEEE/CVF conference on computer vision and pattern recognition}, pages 4401--4410, 2019.

\bibitem[Ke et~al.(2021)Ke, Wang, Wang, Milanfar, and Yang]{ke2021musiq}
Junjie Ke, Qifei Wang, Yilin Wang, Peyman Milanfar, and Feng Yang.
\newblock Musiq: Multi-scale image quality transformer.
\newblock In \emph{Proceedings of the IEEE/CVF International Conference on Computer Vision}, pages 5148--5157, 2021.

\bibitem[Kingma and Welling(2013)]{kingma2013vae}
Diederik~P Kingma and Max Welling.
\newblock Auto-encoding variational bayes.
\newblock \emph{arXiv preprint arXiv:1312.6114}, 2013.

\bibitem[Lee et~al.(2013)Lee, Lee, and Kim]{lee2013dicm}
Chulwoo Lee, Chul Lee, and Chang-Su Kim.
\newblock Contrast enhancement based on layered difference representation of 2d histograms.
\newblock \emph{IEEE transactions on image processing}, 22\penalty0 (12):\penalty0 5372--5384, 2013.

\bibitem[Li et~al.(2021)Li, Guo, and Loy]{UnSupervised_li2021zerodcepp}
Chongyi Li, Chunle Guo, and Chen~Change Loy.
\newblock Learning to enhance low-light image via zero-reference deep curve estimation.
\newblock \emph{IEEE Transactions on Pattern Analysis and Machine Intelligence}, 44\penalty0 (8):\penalty0 4225--4238, 2021.

\bibitem[Liang et~al.(2022{\natexlab{a}})Liang, Li, Wei, Yang, Zhang, Yang, Du, and Zhou]{UnSupervised_liang2022scllle}
Dong Liang, Ling Li, Mingqiang Wei, Shuo Yang, Liyan Zhang, Wenhan Yang, Yun Du, and Huiyu Zhou.
\newblock Semantically contrastive learning for low-light image enhancement.
\newblock In \emph{Proceedings of the AAAI Conference on Artificial Intelligence}, pages 1555--1563, 2022{\natexlab{a}}.

\bibitem[Liang et~al.(2022{\natexlab{b}})Liang, Xu, Quan, Shi, and Ji]{Zeroshot_liang2022dunp}
Jinxiu Liang, Yong Xu, Yuhui Quan, Boxin Shi, and Hui Ji.
\newblock Self-supervised low-light image enhancement using discrepant untrained network priors.
\newblock \emph{IEEE Transactions on Circuits and Systems for Video Technology}, 32\penalty0 (11):\penalty0 7332--7345, 2022{\natexlab{b}}.

\bibitem[Liang et~al.(2023)Liang, Li, Zhou, Feng, and Loy]{UnSupervised_liang2023iterative}
Zhexin Liang, Chongyi Li, Shangchen Zhou, Ruicheng Feng, and Chen~Change Loy.
\newblock Iterative prompt learning for unsupervised backlit image enhancement.
\newblock In \emph{Proceedings of the IEEE/CVF International Conference on Computer Vision}, pages 8094--8103, 2023.

\bibitem[Liu et~al.(2021)Liu, Ma, Zhang, Fan, and Luo]{UnSupervised_liu2021ruas}
Risheng Liu, Long Ma, Jiaao Zhang, Xin Fan, and Zhongxuan Luo.
\newblock Retinex-inspired unrolling with cooperative prior architecture search for low-light image enhancement.
\newblock In \emph{Proceedings of the IEEE/CVF Conference on Computer Vision and Pattern Recognition}, pages 10561--10570, 2021.

\bibitem[Liu et~al.(2016)Liu, Anguelov, Erhan, Szegedy, Reed, Fu, and Berg]{Degradation_liu2016ssd}
Wei Liu, Dragomir Anguelov, Dumitru Erhan, Christian Szegedy, Scott Reed, Cheng-Yang Fu, and Alexander~C Berg.
\newblock Ssd: Single shot multibox detector.
\newblock In \emph{Computer Vision--ECCV 2016: 14th European Conference, Amsterdam, The Netherlands, October 11--14, 2016, Proceedings, Part I 14}, pages 21--37. Springer, 2016.

\bibitem[Ma et~al.(2015)Ma, Zeng, and Wang]{ma2015mef17}
Kede Ma, Kai Zeng, and Zhou Wang.
\newblock Perceptual quality assessment for multi-exposure image fusion.
\newblock \emph{IEEE Transactions on Image Processing}, 24\penalty0 (11):\penalty0 3345--3356, 2015.

\bibitem[Ma et~al.(2022)Ma, Ma, Liu, Fan, and Luo]{UnSupervised_ma2022sci}
Long Ma, Tengyu Ma, Risheng Liu, Xin Fan, and Zhongxuan Luo.
\newblock Toward fast, flexible, and robust low-light image enhancement.
\newblock In \emph{Proceedings of the IEEE/CVF Conference on Computer Vision and Pattern Recognition}, pages 5637--5646, 2022.

\bibitem[Mittal et~al.(2012)Mittal, Soundararajan, and Bovik]{mittal2012niqe}
Anish Mittal, Rajiv Soundararajan, and Alan~C Bovik.
\newblock Making a “completely blind” image quality analyzer.
\newblock \emph{IEEE Signal processing letters}, 20\penalty0 (3):\penalty0 209--212, 2012.

\bibitem[Ni et~al.(2020)Ni, Yang, Wang, Ma, and Kwong]{UnSupervised_ni2020uegan}
Zhangkai Ni, Wenhan Yang, Shiqi Wang, Lin Ma, and Sam Kwong.
\newblock Towards unsupervised deep image enhancement with generative adversarial network.
\newblock \emph{IEEE Transactions on Image Processing}, 29:\penalty0 9140--9151, 2020.

\bibitem[Radford et~al.(2015)Radford, Metz, and Chintala]{radford2015dcgan}
Alec Radford, Luke Metz, and Soumith Chintala.
\newblock Unsupervised representation learning with deep convolutional generative adversarial networks.
\newblock \emph{arXiv preprint arXiv:1511.06434}, 2015.

\bibitem[Razavi et~al.(2019)Razavi, Van~den Oord, and Vinyals]{razavi2019vqvae2}
Ali Razavi, Aaron Van~den Oord, and Oriol Vinyals.
\newblock Generating diverse high-fidelity images with vq-vae-2.
\newblock \emph{Advances in neural information processing systems}, 32, 2019.

\bibitem[Rombach et~al.(2022)Rombach, Blattmann, Lorenz, Esser, and Ommer]{rombach2022latentdiff}
Robin Rombach, Andreas Blattmann, Dominik Lorenz, Patrick Esser, and Bj{\"o}rn Ommer.
\newblock High-resolution image synthesis with latent diffusion models.
\newblock In \emph{Proceedings of the IEEE/CVF conference on computer vision and pattern recognition}, pages 10684--10695, 2022.

\bibitem[Ulyanov et~al.(2018)Ulyanov, Vedaldi, and Lempitsky]{ulyanov2018dip}
Dmitry Ulyanov, Andrea Vedaldi, and Victor Lempitsky.
\newblock Deep image prior.
\newblock In \emph{Proceedings of the IEEE conference on computer vision and pattern recognition}, pages 9446--9454, 2018.

\bibitem[Van Den~Oord et~al.(2017)Van Den~Oord, Vinyals, et~al.]{van2017vqvae}
Aaron Van Den~Oord, Oriol Vinyals, et~al.
\newblock Neural discrete representation learning.
\newblock \emph{Advances in neural information processing systems}, 30, 2017.

\bibitem[Wang et~al.(2013)Wang, Zheng, Hu, and Li]{wang2013npe}
Shuhang Wang, Jin Zheng, Hai-Miao Hu, and Bo Li.
\newblock Naturalness preserved enhancement algorithm for non-uniform illumination images.
\newblock \emph{IEEE transactions on image processing}, 22\penalty0 (9):\penalty0 3538--3548, 2013.

\bibitem[Wang et~al.(2004)Wang, Bovik, Sheikh, and Simoncelli]{wang2004ssim}
Zhou Wang, Alan~C Bovik, Hamid~R Sheikh, and Eero~P Simoncelli.
\newblock Image quality assessment: from error visibility to structural similarity.
\newblock \emph{IEEE transactions on image processing}, 13\penalty0 (4):\penalty0 600--612, 2004.

\bibitem[Wei et~al.(2018)Wei, Wang, Yang, and Liu]{Supervised_2018Retinexnet}
Chen Wei, Wenjing Wang, Wenhan Yang, and Jiaying Liu.
\newblock Deep retinex decomposition for low-light enhancement.
\newblock \emph{arXiv preprint arXiv:1808.04560}, 2018.

\bibitem[Wu et~al.(2023)Wu, Duan, Guo, Chai, and Li]{wu2023ridcp}
Rui-Qi Wu, Zheng-Peng Duan, Chun-Le Guo, Zhi Chai, and Chongyi Li.
\newblock Ridcp: Revitalizing real image dehazing via high-quality codebook priors.
\newblock In \emph{Proceedings of the IEEE/CVF Conference on Computer Vision and Pattern Recognition}, pages 22282--22291, 2023.

\bibitem[Wu et~al.(2022)Wu, Weng, Zhang, Wang, Yang, and Jiang]{Supervised_wu2022uRetinex}
Wenhui Wu, Jian Weng, Pingping Zhang, Xu Wang, Wenhan Yang, and Jianmin Jiang.
\newblock Uretinex-net: Retinex-based deep unfolding network for low-light image enhancement.
\newblock In \emph{Proceedings of the IEEE/CVF conference on computer vision and pattern recognition}, pages 5901--5910, 2022.

\bibitem[Xu et~al.(2022)Xu, Wang, Fu, and Jia]{Supervised_xu2022snr}
Xiaogang Xu, Ruixing Wang, Chi-Wing Fu, and Jiaya Jia.
\newblock Snr-aware low-light image enhancement.
\newblock In \emph{Proceedings of the IEEE/CVF conference on computer vision and pattern recognition}, pages 17714--17724, 2022.

\bibitem[Yang et~al.(2023)Yang, Ding, Wu, Li, and Zhang]{UnSupervised_yang2023nerco}
Shuzhou Yang, Moxuan Ding, Yanmin Wu, Zihan Li, and Jian Zhang.
\newblock Implicit neural representation for cooperative low-light image enhancement.
\newblock In \emph{Proceedings of the IEEE/CVF International Conference on Computer Vision}, pages 12918--12927, 2023.

\bibitem[Ying et~al.(2022)Ying, He, Gao, Han, and Chu]{ying2022eagan}
Guohao Ying, Xin He, Bin Gao, Bo Han, and Xiaowen Chu.
\newblock Eagan: Efficient two-stage evolutionary architecture search for gans.
\newblock In \emph{European Conference on Computer Vision}, pages 37--53. Springer, 2022.

\bibitem[Zhang et~al.(2019)Zhang, Zhang, and Guo]{Supervised_zhang2019kind}
Yonghua Zhang, Jiawan Zhang, and Xiaojie Guo.
\newblock Kindling the darkness: A practical low-light image enhancer.
\newblock In \emph{Proceedings of the 27th ACM international conference on multimedia}, pages 1632--1640, 2019.

\bibitem[Zhang et~al.(2021)Zhang, Guo, Ma, Liu, and Zhang]{Supervised_zhang2021kindpp}
Yonghua Zhang, Xiaojie Guo, Jiayi Ma, Wei Liu, and Jiawan Zhang.
\newblock Beyond brightening low-light images.
\newblock \emph{International Journal of Computer Vision}, 129:\penalty0 1013--1037, 2021.

\bibitem[Zheng et~al.(2021)Zheng, Shi, and Shi]{Degradation_Zheng_2021_ICCV}
Chuanjun Zheng, Daming Shi, and Wentian Shi.
\newblock Adaptive unfolding total variation network for low-light image enhancement.
\newblock pages 4439--4448, 2021.

\bibitem[Zheng and Gupta(2022)]{UnSupervised_zheng2022SGZ}
Shen Zheng and Gaurav Gupta.
\newblock Semantic-guided zero-shot learning for low-light image/video enhancement.
\newblock In \emph{Proceedings of the IEEE/CVF Winter conference on applications of computer vision}, pages 581--590, 2022.

\bibitem[Zhou et~al.(2022)Zhou, Chan, Li, and Loy]{zhou2022codeformer}
Shangchen Zhou, Kelvin Chan, Chongyi Li, and Chen~Change Loy.
\newblock Towards robust blind face restoration with codebook lookup transformer.
\newblock \emph{Advances in Neural Information Processing Systems}, 35:\penalty0 30599--30611, 2022.

\end{thebibliography}
